\documentclass[final,1p,times,authoryear]{elsarticle}

\usepackage{amsmath,amssymb,amsfonts}
\usepackage{algorithmic}
\usepackage{graphicx}
\usepackage{textcomp}
\usepackage{xcolor}

\usepackage{url}

\usepackage[small]{caption}
\usepackage[hidelinks]{hyperref}
\usepackage{algorithm}
\usepackage{algorithmic}

\usepackage[utf8]{inputenc}
\usepackage{booktabs}
\usepackage{wrapfig}
\usepackage{microtype}

\usepackage{amsfonts}
\usepackage{blindtext}
\usepackage{microtype}
\usepackage{mathrsfs}

\usepackage[USenglish]{babel}
\usepackage{csquotes}
\usepackage{multirow,adjustbox}
\usepackage{subcaption}

\newcommand{\maml}{\textsc{maml}}
\newcommand{\reptile}{\textsc{reptile}}
\newcommand{\adam}{\textsc{adam}}

\DeclareMathOperator{\enc}{enc}
\newcommand {\train}  {{\text{train}}}
\newcommand {\test}   {{\text{test}}}

\newcommand {\init}   {{\text{init}}}

\newcommand {\R}   {{\mathbb R}}

\newcommand {\T}   {{\mathcal T}}

\DeclareMathOperator* {\argmin}   {arg min}

\newcommand{\cha}{\textsc{chameleon}}

\newcommand{\ordma}{reordering}

\newcommand{\rpm}{\raisebox{.2ex}{$\scriptstyle\pm$}}

\journal{Neural Networks}
\begin{document}

\begin{frontmatter}

\title{Chameleon: Learning Model Initializations Across Tasks With Different Schemas}

\author[ismll,eq] {Lukas Brinkmeyer}
\author[ismll,eq] {Rafael Rego Drumond}
\author[ismll] {Randolf Scholz}
\author[noti]  {Josif Grabocka}
\author[ismll] {Lars Schmidt-Thieme}

\address[eq]   {Equal Contribution}
\address[ismll]{University of Hildesheim, Information Systems and
Machine Learning Lab, Hildesheim, Germany (\{brinkmeyer,radrumond,scholz,schmidt-thieme\}@ismll.uni-hildehseim.de)}
\address[noti] {Albert-Ludwigs-Universität Freiburg, Freiburg, Germany (grabocka@informatik.uni-freiburg.de)}
\begin{abstract}

Parametric models, and particularly neural networks, require weight initialization as a starting point for gradient-based optimization. Recent work shows that a specific initial parameter set can be learned from a population of supervised learning tasks. Using this initial parameter set enables a fast convergence for unseen classes even when only a handful of instances is available (model-agnostic meta-learning). 
Currently, methods for learning model initializations are limited to a population of tasks sharing the same schema, i.e., the same number, order, type, and semantics of predictor and target variables.
In this paper, we address the problem of meta-learning parameter initialization across tasks with different schemas, i.e., if the number of predictors varies across tasks, while they still share some variables. We propose Chameleon, a model that learns to align different predictor schemas to a common representation. 
In experiments on 23 datasets of the OpenML-CC18 benchmark, we show that Chameleon can successfully learn parameter initializations across tasks with different schemas, presenting, to the best of our knowledge, the first cross-dataset few-shot classification approach for unstructured data.

\end{abstract}

\end{frontmatter}
\section{Introduction}

    \begin{figure}[t!]
        \centering
        \includegraphics[width=1\textwidth]{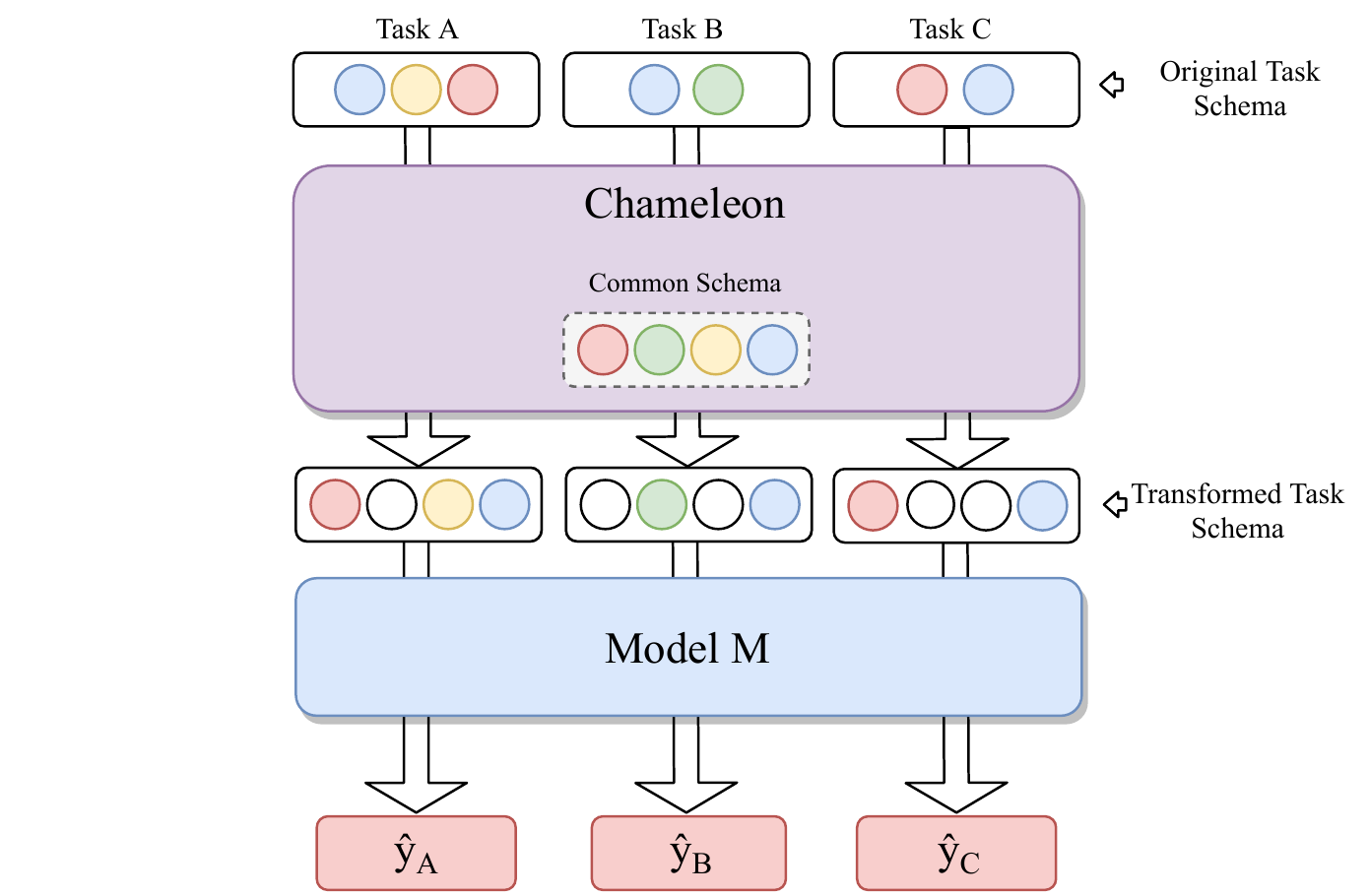}
        \caption{\textbf{Chameleon Pipeline}: The main idea of Chameleon is to encode tasks with different schemas to a shared representation with a uniform feature space, which can then be processed by any classifier.
            In this picture, the top part represents tasks of the same domain with different schemas. The bottom part represents the aligned features in a fixed schema.
            } \label{tli}
    \end{figure}
    Humans require only a few examples to correctly classify new instances of previously unknown objects. For example, it is sufficient to see a handful of images of a specific type of dog before being able to classify dogs of this type consistently. In contrast, deep learning models optimized in a classical supervised setup usually require a vast number of training examples to match human performance. A striking difference is that a human has already learned to classify countless other objects, while parameters of a neural network are typically initialized randomly. Previous approaches improved this starting point for gradient-based optimization by choosing a more robust random initialization \citep{He_2015_ICCV} or by starting from a pretrained network \citep{pan2010survey}. Still, they failed to enable learning from only a handful of training examples. Moreover, established hyperparameter optimization methods \citep{schilling2016scalable} are not capable of optimizing the model initialization due to the high-dimensional parameter space.

    Few-shot classification aims at correctly classifying unseen instances of a novel task with only a few labeled training instances given. This is typically accomplished by meta-learning across a set of training tasks, which consist of training and validation examples with given labels for a set of classes.
    The field has gained immense popularity among researchers after recent meta-learning approaches have shown that it is possible to optimize a single model across different tasks, which can classify novel classes after seeing only a few instances \citep{finn2018one}. However, training a single model across different tasks is only feasible if all tasks share the same schema, meaning that all instances share one set of features in identical order. For that reason, most current approaches demonstrate their performance on image data, which can be easily scaled to a fixed shape, whereas transforming unstructured data to a uniform schema is not trivial.
    
    We want to extend popular approaches to operate schema invariant, i.e., independent of order and shape, making it possible to use meta-learning approaches on unstructured data with varying feature spaces, e.g., learning a model from hearts disease data that can accurately classify a few-shot task for diabetes detection that relies on similar features. Thus, we require a schema-invariant encoder that maps hearts and diabetes data to one feature representation, which then can be used to train a single model via popular meta-learning algorithms like \reptile{} \citep{reptile2018}.
    
    We propose a set-wise feature transformation model called \cha{}, named after a \reptile{} capable of adjusting its colors according to the environment in which it is located. \cha{} deals with different schemas by projecting them to a fixed input space while keeping features from different tasks but of the same type or distribution in the same position, as illustrated by Figure \ref{tli}. Our model learns to compute a task-specific reordering matrix that, when multiplied with the original input, aligns the schema of unstructured tasks to a common representation while behaving invariant to the order of input features.
    
    Our main contributions are as follows:
    (1) We show how our proposed method \cha{} can learn to align varying feature spaces to a common representation.
    (2) We propose the first approach to tackle few-shot classification for tasks with different schemas.
    (3) In experiments on 23 datasets of the OpenML-CC18 benchmark \citep{bischl2017openml} collection, we demonstrate how current meta-learning approaches can successfully learn a model initialization across tasks with different schemas as long as they share some variables with respect to their type or semantics. 
    (4) Although an alignment makes little sense to be performed on top of structured data such as images that can be easily rescaled, we demonstrate how \cha{} can align latent embeddings of two image datasets generated with different neural networks. \section{Related Work} \label{related}
    
    Our goal is to extend recent few-shot classification approaches that make use of optimization-based meta-learning by adding a feature alignment component that casts different inputs to a common schema, presenting the first approach working across tasks with different schema. In this section, we will discuss various works related to our approach.

    Research on transfer learning \citep{pan2010survey,sung2018comp,gligic2020named} has shown that training a model on different auxiliary tasks before actually fitting it to the target problem can provide better results if training data is scarce. Motivated by this, few-shot learning approaches try to generalize to unseen classes given only a few instances \citep{duan2017one,finn2017one,snell2017prototypical}. Several meta-learning approaches have been developed to tackle this problem by introducing architectures and parameterization techniques specifically suited for few-shot classification \citep{mishra2018snail,SHI201947,WANG20201} while others try to extract useful meta-features from datasets to improve hyper-parameter optimization \citep{jomaa2019dataset2vec}. All of these approaches operate on a set of tasks. A task $\tau\in\T$ consists of predictor data $X_\tau$, a target $Y_\tau$, a predefined training/test split ${\tau = (X^\train_\tau, Y^\train_\tau, X^\test_\tau, Y^\test_\tau)}$ and a loss function $L_\tau$.
    
    Moreover, optimization-based approaches, as proposed by \cite{finn2017model} show that an adapted learning paradigm can be sufficient for learning across tasks.
    Model Agnostic Meta-Learning (\maml{}) describes a model initialization algorithm that is capable of training an arbitrary model $f$ across different tasks. Instead of sequentially training the model one task at a time, it uses update steps from different tasks to find a common gradient direction that achieves a fast convergence for these.
    In other words, for each meta-learning update, we would need an initial value for the model parameters $\theta$. Then, we sample a batch of tasks $\T$, and for each task $\tau \in \T$ we find an updated version of $\theta$ using $N$ examples from the task performing gradient descent as in:
    \begin{equation}
        \theta'_i \leftarrow \theta - \alpha \nabla_\theta \mathcal{L}_{\tau_i}(f_\theta)        
    \end{equation}
    The final update of $\theta$ will be:%
    \begin{equation} \label{ma2}
        \theta \leftarrow \theta - \beta \frac{1}{|\T|} \nabla_\theta \sum_{i} \mathcal{L}_{\tau_i}(f_{\theta'_i})        
    \end{equation}    
    \cite{finn2017model} state that \maml{} does not require learning an update rule \citep{ravi2016optimization}, or restricting their model architecture \citep{santoro2016meta}. They extended their approach by incorporating a probabilistic component such that for a new task, the model is sampled from a distribution of models to guarantee a higher model diversification for ambiguous tasks \citep{finn2018one}.
    However, \maml{} requires to compute second order derivatives, ending up being computationally heavy.
    \cite{reptile2018} simplified \maml{} and presented \reptile{} by numerically approximating Equation (\ref{ma2}) to replace the second derivative with the weights difference:
    \begin{equation} \label{rep}
        \theta \leftarrow \theta - \beta   \frac{1}{|\T|} \sum_{i} (\theta'_i - \theta)        
    \end{equation}
    Which means we can use the difference between the previous and updated version as an approximation of the second derivatives to reduce computational cost. The serial version is presented in Algorithm~(1).%

    All of these approaches rely on a fixed schema i.e. the same set of features with identical alignment across all tasks. However, many similar datasets only share a subset of their features, while oftentimes having a different order or representation.  Moreover, most current few-shot classification approaches sample tasks from a single dataset by selecting a random subset of classes and using a train/test split for these although it is possible to train a single meta-model on two different image datasets as shown by \cite{munk2017meta}. Recently, a meta-dataset for few-shot classification of image tasks was also published to promote meta-learning across multiple datasets \citep{Triantafillou2020Meta-Dataset}. 
    Optimizing a single model across various datasets requires a uniform feature space. Thus, it is required to align the features which is achieved by simply rescaling all instances in the case of image data but not trivial for unstructured data. Furthermore, recent work relies on preprocessing the images to a one-dimensional latent embedding with an additional deep neural network \citep{leo} in which case scaling is only applicable beforehand.
    
    \begin{algorithm}[H] \label{alg1}
        \caption{\reptile{} \cite{reptile2018}}
        \textbf{Input}: Meta-dataset $\T$, learning rate $\beta$ 

        \begin{algorithmic}[1]
            \STATE Randomly initialize model $f$ with weight parameters $\theta$
             
             \FOR{iteration = 1, 2, ...}
             
                \STATE Sample task $\tau\sim\T$ with loss $L_\tau$
                
                \STATE $\theta' \leftarrow \theta$
                
                \FOR{k steps = 1,2,...}
                
                    \STATE $\theta' \leftarrow \theta' - \alpha \nabla_{\theta'} L_{\tau}(Y_\tau, f(X_\tau; \theta'))$
                
                \ENDFOR
                
                \STATE $\theta \leftarrow \theta - \beta (\theta' - \theta)$  
                
             \ENDFOR
             
             \STATE \textbf{return} model $f$ , weight parameters $\theta$
            % \hline
            %  \hline
             
        \end{algorithmic}
    \end{algorithm}
    
    Few approaches in the literature deal with feature alignment. The work from \cite{pan2010Sen} describes a procedure for sentiment analysis that aligns words across different domains that have the same sentiment. Recently, \cite{zhang2018local} proposed an unsupervised framework called Local Deep-Feature Alignment. The procedure computes the global alignment, which is shared across all local neighborhoods by multiplying the local representations with a transformation matrix. So far, none of these methods find a common alignment between features of different datasets.
    We propose a novel feature alignment component named \cha{}, which enables state-of-the-art methods to work on top of tasks whose feature vector differ not only in their length but also their concrete alignment.
\section{Methodology}\label{meth} %

Methods like \reptile{} and \maml{} try to find the best initialization for a specific model, in this work referred to as $\hat{y}$, to work on a set $\T$ of similar tasks. 
However, every task $\tau$ has to share the same schema of common size $K$, where similar features shared across tasks are in the same position.
A feature-order invariant encoder is needed to map the data representation $X_\tau$ of tasks with varying input schema and feature length $F_\tau$ to a shared latent representation $\widetilde{X_\tau}$ with fixed feature length $K$.

\begin{equation}
  \enc \colon
  X_\tau \in\R^{N \times F_\tau} \longmapsto \widetilde{X}_\tau \in \R^{N \times K}
\end{equation}
Where $N$ represents the number of instances in $X_\tau$, $F_\tau$ is the number of features of task $\tau$, and $K$ is the size of the desired feature space. By combining this encoder with a model $\hat y$ that works on a fixed input size $K$ and outputs the predicted target e.g. binary classification, it is possible to apply the \reptile{} algorithm to learn an initialization $\theta^\init$ across tasks with different schema. The optimization objective then becomes the meta-loss for the combined network ${f=\hat{y}\circ\enc}$ over a set of tasks $\T$:
\begin{align}\label{eq: encoder meta loss}
\begin{split}
  \argmin_{\theta^\init}& \quad \mathbb E_{\tau \sim \T} 
  L_\tau\Big(Y_\tau^\test,\; f\big(X_\tau^\test; \theta^{(u)}_\tau\big)\Big) \\
    \text{s.t.}& \quad \theta^{(u)}_\tau = \mathcal{A}^{(u)}\Big(X^\train_\tau, Y^\train_\tau, L_\tau, f; \theta^\init \Big)  
\end{split}
\end{align}
Where ${\theta}^{\init}$ is the concatenation of the initial weights $\theta^{\init}_{\enc}$ for the encoder and $\theta^{\init}_{\hat{y}}$ for model $\hat{y}$, and $\theta^{(u)}_{\tau}$ are the updated weights after applying the learning procedure $\mathcal{A}$ for $u$ iterations on the task $\tau$ as defined in Algorithm 1 for the inner updates of \reptile{}.

\begin{figure}[t!]
\centering
\includegraphics[width=0.8\columnwidth]{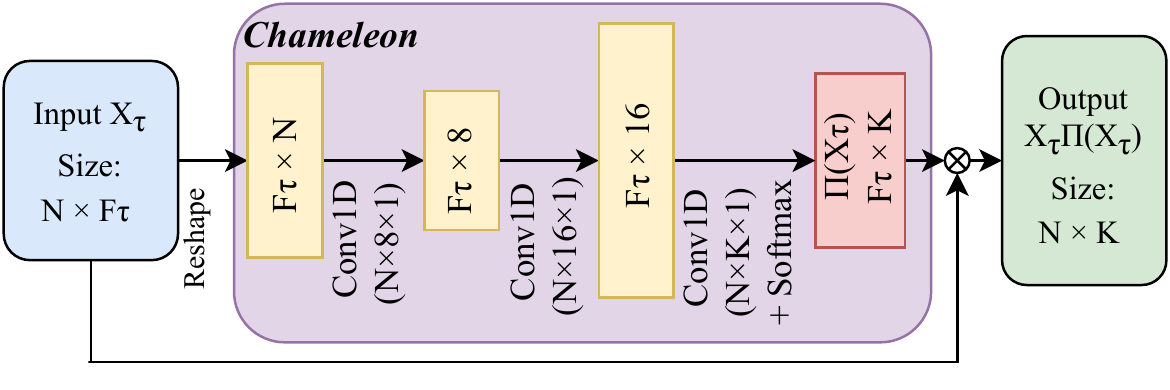}
\caption{\textbf{The Chameleon Architecture:} $N$ represents the number of samples in $\tau$, $F_\tau$ is equal to the feature length of $\tau$, and K is defined as the desired feature space. ``Conv($a\times b\times c$)" is a convolution operation with $a$ input channels, filter size of $b$ and kernel length~$c$.
} \label{chacomp}
\end{figure}

It is important to mention that learning one weight parameterization across any heterogeneous set of tasks is extremely difficult since it is most likely impossible to find one initialization for two tasks with a vastly different number and type of features. By contrast, if two tasks share similar features, one can align the similar features to a common representation so that a model can directly learn across different tasks by transforming the tasks as illustrated in Figure \ref{tli}.

\subsection{Chameleon}
Consider a set of tasks where a \ordma{}  matrix $\Pi_\tau$ exists for each task that transforms predictor data $X_\tau$ into $\widetilde{X}_\tau$ with $\widetilde{X}_\tau$ having the same schema for every task $\tau$:

\begin{align}\label{matmul}
\widetilde{X}_\tau  = X_\tau \cdot \Pi_\tau, \text{where}
\end{align} 
\begin{equation*}
\underbrace{
\begin{bmatrix}
    \tilde x_{1,1} & \dots  & \tilde x_{1,K} \\
    \vdots  & \ddots & \vdots  \\
    \tilde x_{N,1} & \dots  & \tilde x_{N,K}
\end{bmatrix}
}_{\widetilde{X}_\tau}
=
\underbrace{
\begin{bmatrix}
    x_{1,1} & \dots  & x_{1,F_\tau} \\
     \vdots & \ddots & \vdots  \\
    x_{N,1} & \dots  & x_{N,F_\tau}
\end{bmatrix}
}_{X_{\tau}}
\cdot
\underbrace{
\begin{bmatrix}
    \pi_{1,1} & \dots  &\pi_{1,K} \\
    \vdots  & \ddots & \vdots  \\
    \pi_{F_\tau,1} & \dots  & \pi_{F_\tau,K} 
\end{bmatrix}
}_{\Pi_\tau}
\end{equation*}

Every $x_{m,n}$ represents the feature $n$ of sample $m$. Every $\pi_{m,n}$ represent how much of feature $m$ (from samples in $X_{\tau}$) should be shifted to position $n$ in the adapted input $\widetilde{X}_\tau$. Finally, every $\tilde x_{m,n}$ represent the new feature $n$ of sample $m$ in $X_{\tau}$ with the adpated shape and size. This can be also expressed as~${\tilde x_{m,n}=\sum^{F_\tau}_{j=1} x_{m,j}\pi_{j,n}}$. 
In order to achieve the same $\widetilde{X}_\tau$ when permuting two features of a task $X_\tau$, we must simply permute the corresponding rows in $\Pi_\tau$ to achieve the same $\widetilde{X}_\tau$. The summation for every row of $\Pi_\tau$ is set to be equal 1 as in $\sum_i \pi_{j,i} = 1$, so that each value in $\Pi_\tau$ simply states how much a feature is shifted to a corresponding position.

For example: Consider that task $a$ has features [\textit{apples, bananas, melons}] and task $b$ features [\textit{lemons, bananas, apples}]. Both can be transformed to the same representation [\textit{apples, lemons, bananas, melons}] by replacing missing features with zeros and reordering them.
This transformation must have the same result for $a$ and $b$ independently of their feature order. 
In a real life scenario, features might come with different names or sometimes their similarity is not clear to the human eye.
Note that a classical autoencoder is not capable of this as it is not invariant to the order of the features.
Our proposed component, denoted by $\Phi$, takes a task and outputs the corresponding \ordma{} matrix:

\begin{align}
     \Phi (X_\tau, \theta_{\enc})  = \hat{\Pi}_\tau
\end{align}

The function $\Phi$ is a neural network parameterized by $\theta_{\enc}$.
It consists of three 1D-convolutions, where the last one is the output layer that estimates the alignment matrix via a softmax activation.
The input is first transposed to size [$F_{\tau} \times  N$] (where N is the number of samples) i.e., each feature is represented by a vector of instances. Each convolution has kernel length~1 (as the order of instances is arbitrary and thus needs to be permutation invariant) and a channel output size of 8, 16, and lastly $K$. 
The result is a reordering matrix displaying the relation of every original feature to each of the $K$ features in the target space. Each of these vectors passes through a softmax layer, computing the ratio of features in $X_\tau$ shifted to each position of $\widetilde{X}_{\tau}$. 
Finally, the reordering matrix can be multiplied with the input to compute the aligned task as defined in Equation (\ref{matmul}).
By using a kernel length of 1 in combination with the final matrix multiplication, the full architecture becomes permutation invariant in the feature dimension. Column-wise permuting the features of an input task leads to the corresponding row-wise permutation of the reordering matrix. Thus, multiplying both matrices results in the same aligned output independent of permutation.
The overall architecture can be seen in Figure~\ref{chacomp}.
The encoder necessary for training \reptile{} across tasks with different predictor vectors by optimizing Equation (5) is then given as:

\begin{align}
     \enc \colon  X_\tau \longmapsto X_\tau \cdot \Phi(X_\tau, \theta_{\enc}) = X_\tau \cdot {\hat\Pi}_\tau
 \end{align}
 
 \subsection{Reordering-Training}
 
Only joint training the network $\hat y\circ\enc$ as described above, will not teach $\Phi$ how to reorder the features to a shared representation. That is why training $\Phi$ specifically with the objective of reordering features is necessary.
In order to do so, it is essential to optimize the model on a reordering objective while using a meta-dataset that contains similar tasks $\tau$, meaning for each task exists a reordering matrix $\Pi_\tau$ that maps to the shared representation. If $\Pi_\tau$ is known beforehand, optimizing Chameleon becomes a simple classification task based on predicting the new position of each feature in $\tau$. Thus, we can minimize the expected reordering loss over the meta-dataset:
\begin{equation}\label{eq: enc cha}
     \theta_{\enc} = \argmin_{\theta_{\enc}} \mathbb E_{\tau \sim \T^\train} L_{\Phi}\Big(\Pi_\tau, \hat{\Pi}_\tau\Big)
\end{equation}
where $L_{\Phi}$ is the softmax cross-entropy loss, $\Pi_\tau$ is the ground-truth (one-hot encoding of the new position for each variable), and $\hat\Pi_\tau$ is the prediction. This training procedure can be seen in Algorithm~(2). The trained \cha{} model can then be used to compute the $\Pi_\tau$ for any unseen task $\tau\in\mathcal{T}$.

 \begin{algorithm}[H] \label{pretrai}
    \caption{Reordering Training}
    \textbf{Input}: Meta-dataset $\T$, latent dimension $K$, learning rate $\gamma$
    
    \begin{algorithmic}[1]
        
        \STATE Randomly initialize weight parameters $\theta_{\enc}$ of the \cha{} model
         
         \FOR{training iteration = 1, 2, ...}
         
            \STATE randomly sample $\tau\sim \T$
            
            \STATE $\theta_{\enc} \longleftarrow \theta_{\enc} - \gamma\nabla L_\Phi(\Pi_\tau, \Phi(X_\tau,\theta_{\enc}))$   
            
         \ENDFOR
        \STATE return $\theta_{\enc}$
    \end{algorithmic}
  \end{algorithm}

After this training procedure, we can use the learned weights as initialization for $\Phi$ before optimizing $\hat y\circ\enc$ with \reptile{} without further using $L_\Phi$.
Experiments show that this procedure improves our results significantly compared to only optimizing the joint meta-loss.

Training the \cha{} component to reorder similar tasks to a shared representation not only requires a meta-dataset but one where the true reordering matrix $\Pi_\tau$ is provided for every task. In application, this means manually matching similar features of different training tasks so that novel tasks can be matched automatically. However, it is possible to sample a broad number of tasks from a single dataset by sampling smaller sub-tasks from it, selecting a random subset of features in arbitrary order for $N$ random instances. Thus, it is not necessary to manually match the features since all these sub-tasks share the same $\hat{\Pi}_\tau$ apart from the respective permutation of the rows as mentioned above as long as a single dataset is used for meta-training.

\section{Experiments}
In this section, we describe our experimental setup and present the results of our approach on different datasets. As described in the previous section, our architecture consists of the encoder consisting of \cha{} denoted by $\enc$ (Figure \ref{chacomp}) and a base model $\hat y$.
In all of our experiments, we measure the performance of a model and its initialization by evaluating the validation data of a task after performing three update steps on the respective training data. All experiments are conducted in two variants: In \textit{Split} experiments, test tasks contain novel features in addition to features seen during meta-training. In contrast, test tasks in \textit{No-Split} experiments only consist of features seen during meta-training. 
For our baseline results, we analyze the performance of the model $\hat y$ with Glorot initialization \citep{glorot2010understanding} (referred to as \textsc{random}), and with an initialization obtained by regular \reptile{} training on tasks padded to a fixed size $K$ as $\hat y$ is not schema invariant (referred to as $\hat y$). Additionally, we repeat the latter experiment in \textit{No-Split} with tasks sampled from the original dataset by masking a subset of the features. This performance can give an upper estimate as it resembles a perfect alignment of tasks with varying schemas and is referred to as \textsc{oracle}. This baseline is omitted for \textit{Split} experiments as no perfect alignment exists when there are unknown features.

In order to evaluate the proposed method, we investigate the combined model $\hat y\circ\enc$  with the initialization for $\enc$ obtained by training \cha{} as defined in Equation (9) before using \reptile{} to jointly optimize $\hat y \circ \enc$. Furthermore, we repeat this experiment in two ablation studies. First, we do not train \cha{} with Equation (9), but only jointly train $\hat y\circ\enc$ with \reptile{} to evaluate the influence of adding additional parameters to the network. Secondly, we use \reptile{} only to update the initialization for the parameters of $\hat y$ while freezing the previously learned parameters of $\enc$ in order to assess the effect of joint-training both network components. These two variants are referred to as \textsc{untrain} and \textsc{frozen}.
All experiments are conducted with the same model architecture. The base model $\hat{y}$ is a feed-forward neural network with two dense hidden layers that have 16 neurons each. \cha{} consists of two 1D-convolutions with 8 and 16 filters respectively and a final convolution that maps the task to the feature-length $K$, as shown in Figure~\ref{chacomp}.

\begin{table}[t!]
\centering

\begin{tabular}{ccccc}
Dataset & Instances &  Features &  Classes & Full Name \\ \midrule
phonem & 5404 & 5 & 2 & phoneme \\
cmc & 1473 & 24 & 3 & cmc \\
vowel & 990 & 27 & 11 & vowel \\
analca & 797 & 21 & 6 & analcatdata-dmft \\
tic & 958 & 27 & 2 & tic-tac-toe \\
bankno & 1372 & 4 & 2 & banknote-authentication \\
wdbc & 569 & 30 & 2 & wdbc \\
diabet & 768 & 8 & 2 & diabetes \\
segmen & 2310 & 16 & 7 & segment \\
MagicT & 19020 & 10 & 2 & MagicTelescope \\
blood & 748 & 4 & 2 & blood-transfusion-service-center \\
wall & 5456 & 24 & 4 & wall-robot-navigation \\
wilt & 4839 & 5 & 2 & wilt \\
pendig & 10992 & 16 & 10 & pendigits \\
Gestur & 9873 & 32 & 5 & GesturePhaseSegmentationProcessed \\
abalon & 4177 & 10 & 3 & abalone \\
jungle & 44819 & 6 & 3 & jungle-chess-2pcs-raw-endgame-complete \\
letter & 20000 & 16 & 26 & letter \\
ilpd & 583 & 11 & 2 & ilpd \\
wine & 6497 & 11 & 5 & wine-quality \\
mfeat & 2000 & 6 & 10  & mfeat-morphological \\
electr & 45312 & 14 & 2 & electricity \\
vehicl & 846 & 18 & 4 & vehicle \\
 \midrule
\end{tabular} 

\caption{Information for the 23 OpenML-CC18 dataset used in this paper.}
\label{dataTab}
\end{table}
\paragraph{Meta-datasets} For our main experiments, we utilize a single dataset as meta-dataset by sampling the training and test tasks from it. This allows us to evaluate our method on different domains without matching related datasets since $\hat{\Pi}_\tau$ is naturally given for a subset of permuted features. Novel features can also be introduced during testing by splitting not only the instances but also the features of a dataset in train and test partition (\textit{Split}).
Training tasks $\tau\in\T^\train$ are then sampled by selecting a random subset of the training features in arbitrary order for $N$ instances. Stratified sampling guarantees that test tasks $\tau\in\T^\test$ contain both features from train and test while sampling the instances from the test set only. We evaluate our approach using the OpenML-CC18 benchmark \citep{bischl2017openml} from which we selected 23 datasets that have less than 33 features and a minimum number of 90 instances per class. 
We limited the number of features because this work seeks to establish a  proof of concept for learning across data with different schemas. In contrast, very high-dimensional data would require tuning a more complex \cha{} architecture.
The details for each dataset are summarized in Table \ref{dataTab}. The features of each dataset are normalized between 0 and 1. The \textit{Split} experiments are limited to the 21 datasets which have more than four features in order to perform a feature split.
For these experiments, 75\% of the instances are used for reordering-training of \cha{} and joint-training of the full architecture, and 25\% for sampling test tasks. We sample 10 training and 10 validation instances per label for a new task, and 16 tasks per meta-batch. The number of classes in a task is given by the number of classes of the respective dataset, as shown in Table \ref{dataTab}.
During the reordering-training phase and the inner updates of reptile, specified in line 6 of Algorithm (1), we use the \adam{} optimizer \citep{adam} with an initial learning rate of 0.0001 and 0.001 respectively. The meta-updates of \reptile{} are carried out with a learning rate $\beta$ of 0.01. The reordering-training phase is run for 4000 epochs. 

For \textit{Split} experiments, we further impose a train test split on the features (20\% of the features are restricted to the test split). When sampling a task, we sample between 40\% and 60\% of the training features, and for test tasks in the \textit{Split} experiment 20\% of these are from the set of test features. For each experimental run, the different variants are tested on the same data split, and we sample 1600 test tasks beforehand, while the training tasks are randomly sampled each epoch. All experiments are repeated five times with different instance and, in the case of \textit{Split}, different feature splits, and the results are averaged.
Our work is built on top of \reptile{}~\citep{reptile2018} but can be used in conjunction with any model-agnostic meta-learning method. We opted to use \reptile{} since it does not require second derivatives, and the code is publicly available \citep{gitreptile} while also being easy to adapt to our problem.

    \begin{figure} [t!]
              \centering
               \includegraphics[width=0.6\textwidth]{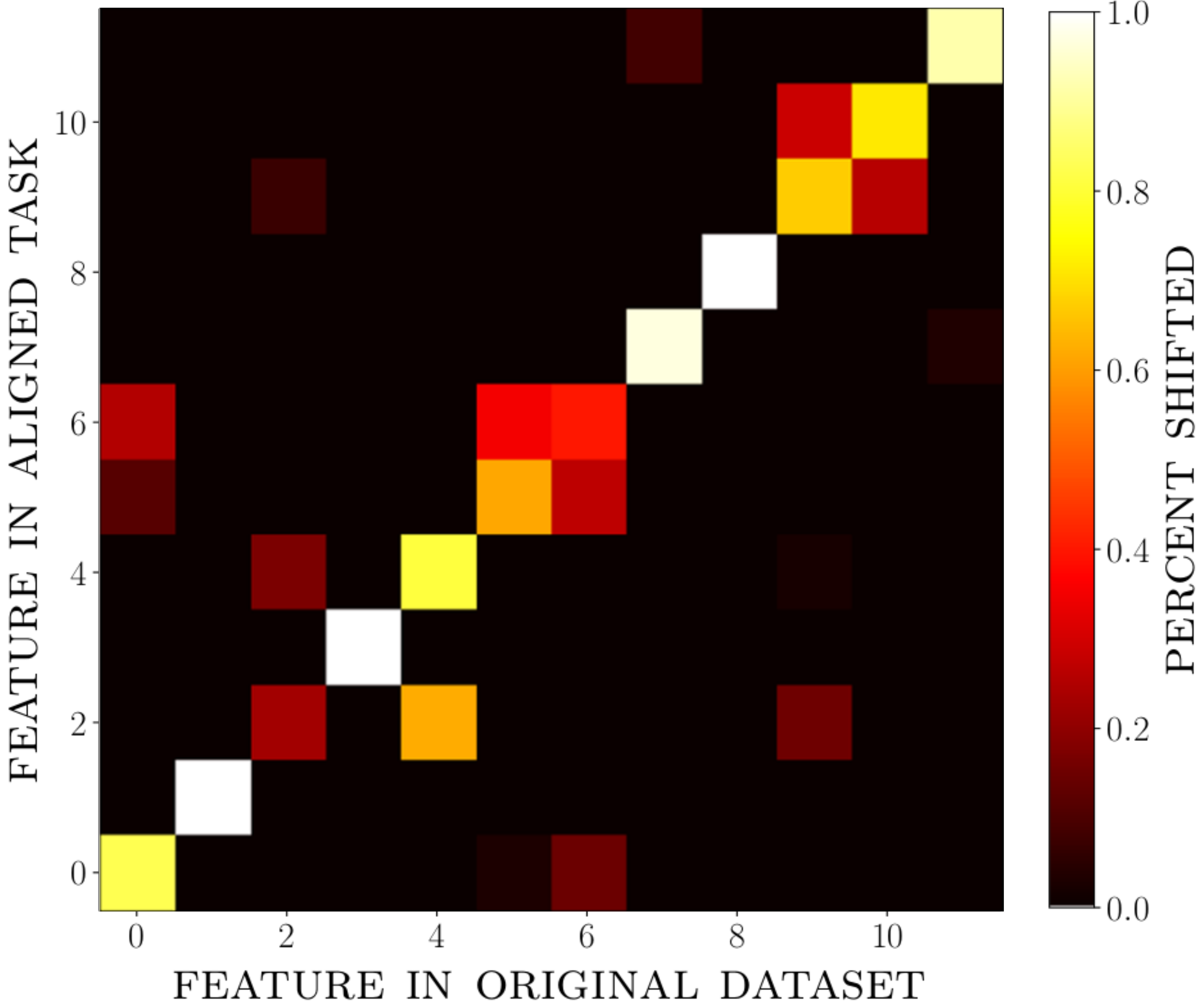}
                \caption{ \textbf{Heat map of the feature shifts for the Wine data computed with \cha{} after reordering-training}: The x-axis represents the twelve features of the original dataset in the correct order and the y-axis shows which position these features are shifted to when presented in a permuted subset. The results are averaged over 1000 sampled tasks.} \label{heat}

    \end{figure}

\paragraph{Experimental results} The result of pretraining \cha{} on the Wine dataset can be seen in Figure \ref{heat}. The x-axis shows the twelve distinct features of the Wine data, and the y-axis represents the predicted position when being presented with a task containing a permuted subset of the features. The color indicates the average portion of the feature that is shifted to the corresponding position. One can see that the component manages to learn the true feature position in almost all cases. Moreover, this illustration does also show that \cha{} can be used to compute the similarity between different features by indicating which pairs are confused most often. For example, features two and three are showing a strong correlation, which is very plausible since they depict the \textit{free sulfur dioxide} and \textit{total sulfur dioxide level} of the wine. This demonstrates that our proposed architecture is able to learn an alignment between different feature spaces~(contribution~1).

    \begin{figure} [t!]
        \centering
        \includegraphics[height=8cm]{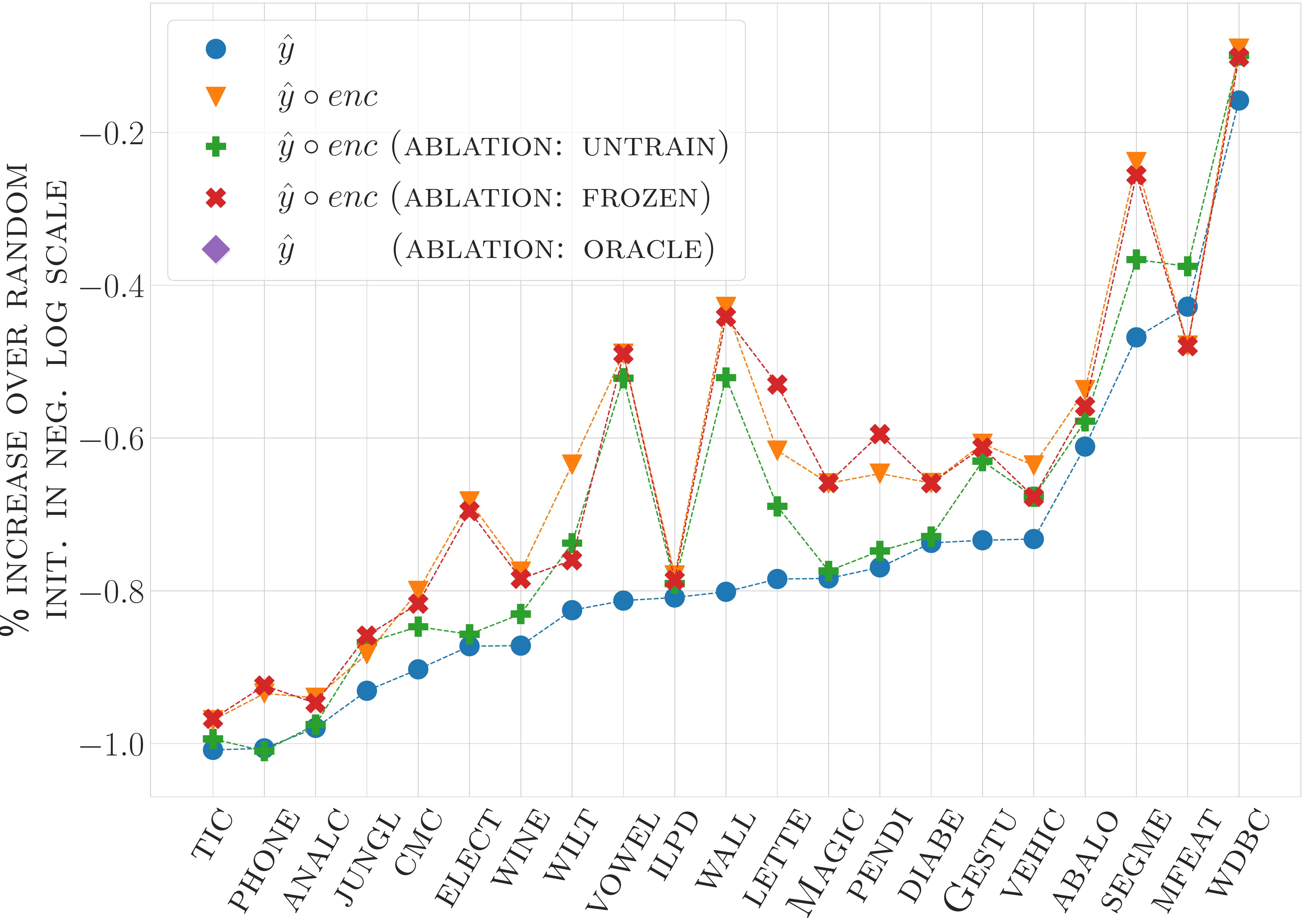} \\
        \includegraphics[clip, trim=0cm 0cm 0cm 0cm, height=8cm]{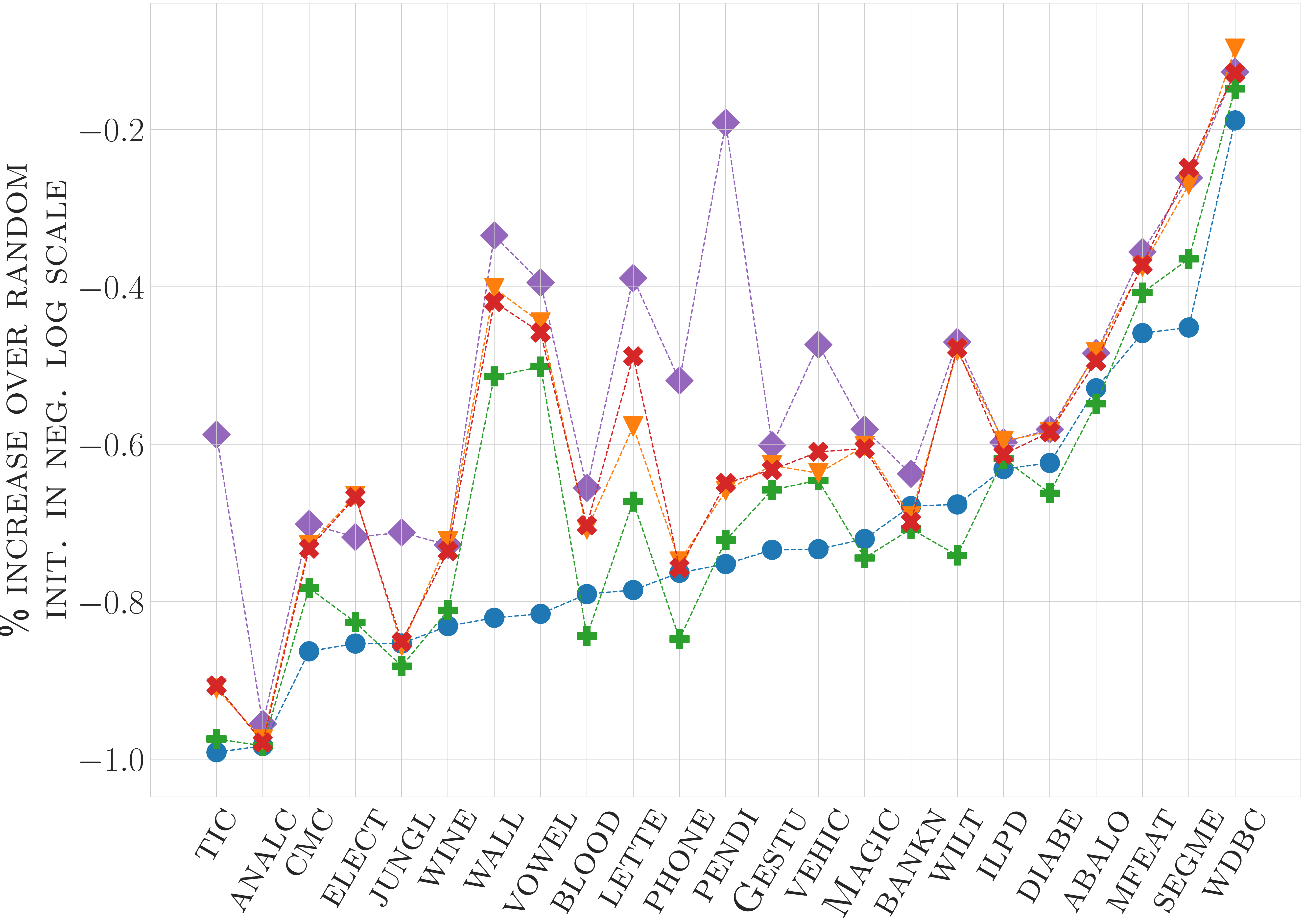}
        \caption{ \textbf{Accuracy improvement for each method over Glorot initialization \citep{glorot2010understanding}}: The difference is plotted in negative log scale to account for the varying performance scales across the different datasets (higher points are better). The top graph represents \textit{Split} experiments while the bottom one depicts the \textit{No-Split} experiments. Notice that the oracle has been omitted from the \textit{Split} experiments since there is no true feature alignment for unseen features. The dataset axis is sorted by the performance of \reptile{} on the base model to improve readability. 
        }\label{rescurves}
    \end{figure}

\begin{figure} [t!]
        \begin{minipage}[t!]{\columnwidth}
           \centering
        \includegraphics[width=.95\textwidth]{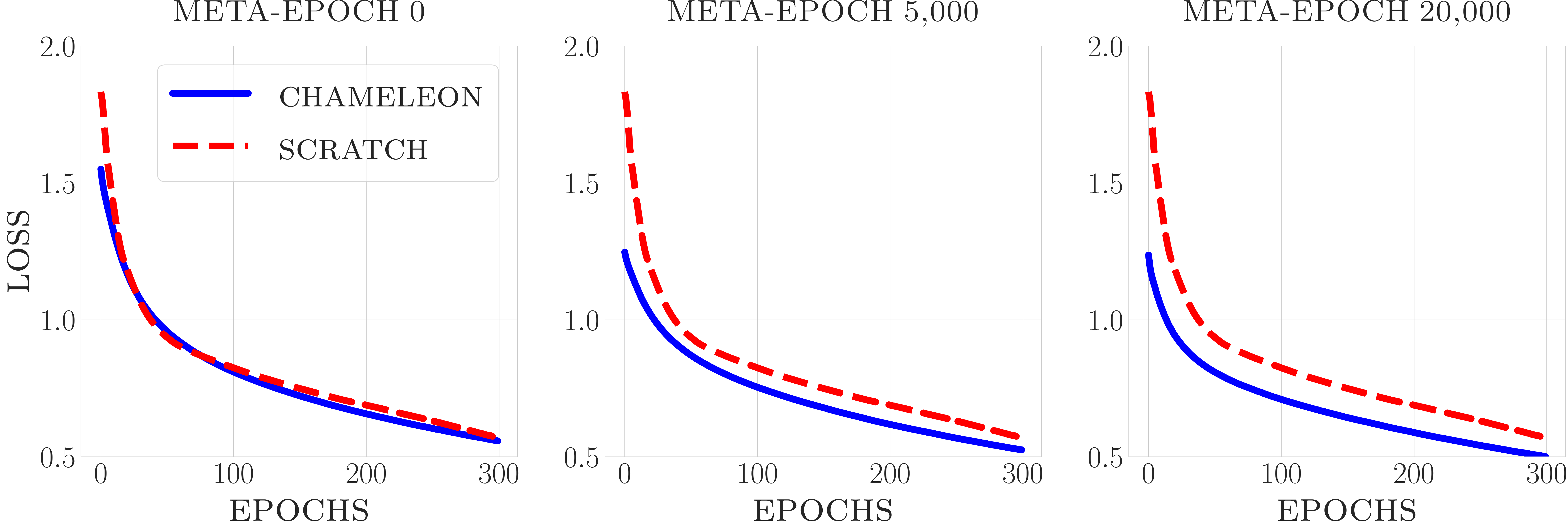}
        \setlength{\belowcaptionskip}{-7pt}
            \caption{\textbf{Snapshots visualizing the inner training}. Validation loss for a task sampled from the \textit{wall-robot-navigation} data set during inner training starting from the current initialization (blue) and random initialization (red). Plots are shown for the initializations obtained from meta-training for 0, 5,000 and 20,000 meta-epochs and trained until convergence. Note that both losses are not identical in meta-epoch 0 because the \cha{} component is already pretrained.
            }
     \label{zoom}
        \end{minipage}   
        
\end{figure}

In Figure \ref{rescurves}, we can see the results for the experiments on the Open-ML benchmark datasets. 
Adding \cha{} without pretraining does not give a significant lift for most datasets. However, adding reordering-training results in a clear performance lift over the other approaches. This demonstrates to the best of our knowledge the first few-shot classification approach, which successfully learns across tasks with varying schemas (contribution 2). The fact that we can only see consistent performance improvements when using reordering-training shows that the feature alignment enables better optimization across tasks with varying schemas. Furthermore, in the \textit{No-Split} results one can see that the performance of the proposed method approaches the \textsc{oracle} performance, which suggests an ideal feature alignment. In all experiments, pretraining \cha{} shows superior performance (contribution 3). When adding novel features during test time \cha{} is still able to outperform the other setups although with a lower margin. The results for \textit{Frozen} \cha{} show that further joint training can improve performance for some datasets, but is not essential in most cases. A detailed overview for these experimental results is given at the end in Tables \ref{statsexpNOSPLIT} and \ref{statsexpSPLIT}.
We visualize the inner training for one of the experiments in Figure \ref{zoom}. It shows three exemplary snapshots of the inner test loss when training on a sampled task with the current initialization $\theta^\init$ during meta-learning. It is compared to the test loss of the model when it is trained on the same task starting with the random initialization. The snapshots show the expected \reptile{} behavior, namely a faster convergence when using the currently learned initialization compared to a random one.
 
\begin{figure} [t!]
        \centering
        \includegraphics[width=.9\columnwidth]{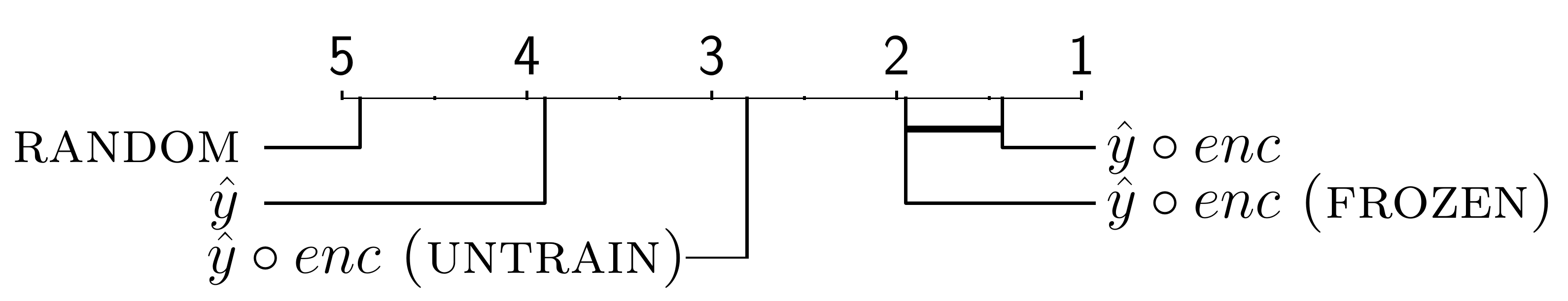}
        \includegraphics[width=.9\columnwidth]{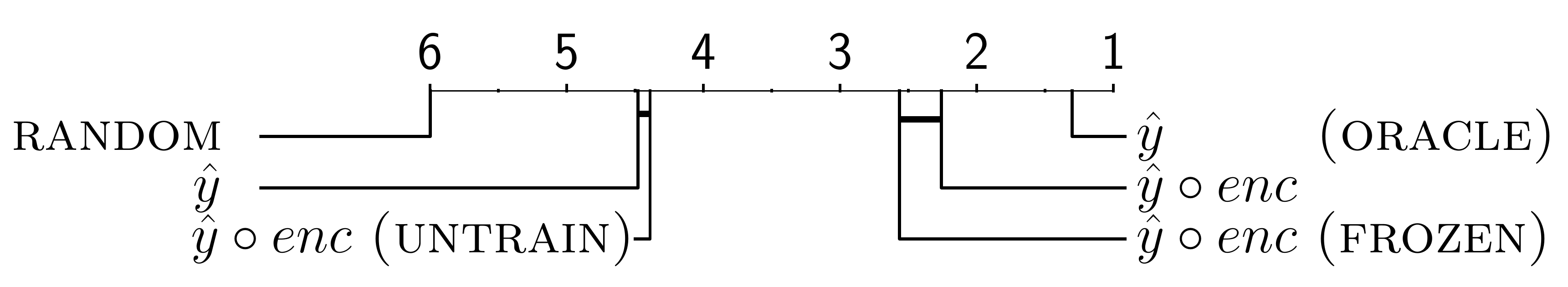}
        \caption{ \textbf{Critical Difference Diagram} for \textit{Split} (Top) and \textit{No-Split} (Bottom) showing results of Wilcoxon signed-rank test with Holm’s alpha correction and 5\% significance level. Models are ranked by their performance and a thicker horizontal line indicates pairs that are not statistically different.}\label{cd-diag}
    \end{figure}

In order to assess the significance of our results, we conducted a Wilcoxon signed-rank test \citep{wilcoxon1992individual} with Holm’s alpha correction \citep{holm1979simple} and displayed the results in the form of a critical difference diagram \citep{demvsar2006statistical} presented in Figure~\ref{cd-diag}. The diagram is generated with the code published by \cite{IsmailFawaz2018deep}. It shows the ranked performance of each model and whether they are statistically different. 
The results confirm that our approach leads to statistically significant improvements over the random and \reptile{} baselines when pretraining \cha{}. Similarly, our approach is also significantly better than jointly training the full architecture without pretraining \cha{} (\textsc{untrain}), confirming that the improvements do not stem from the increased model capacity. Finally, comparing the results to the \textsc{frozen} model shows improvements that are not significant, indicating that a near-optimal alignment was already found during pretraining.

\begin{figure} [t!]
        \begin{minipage}[t!]{\columnwidth}
           \centering
          \includegraphics[width=.9\textwidth]{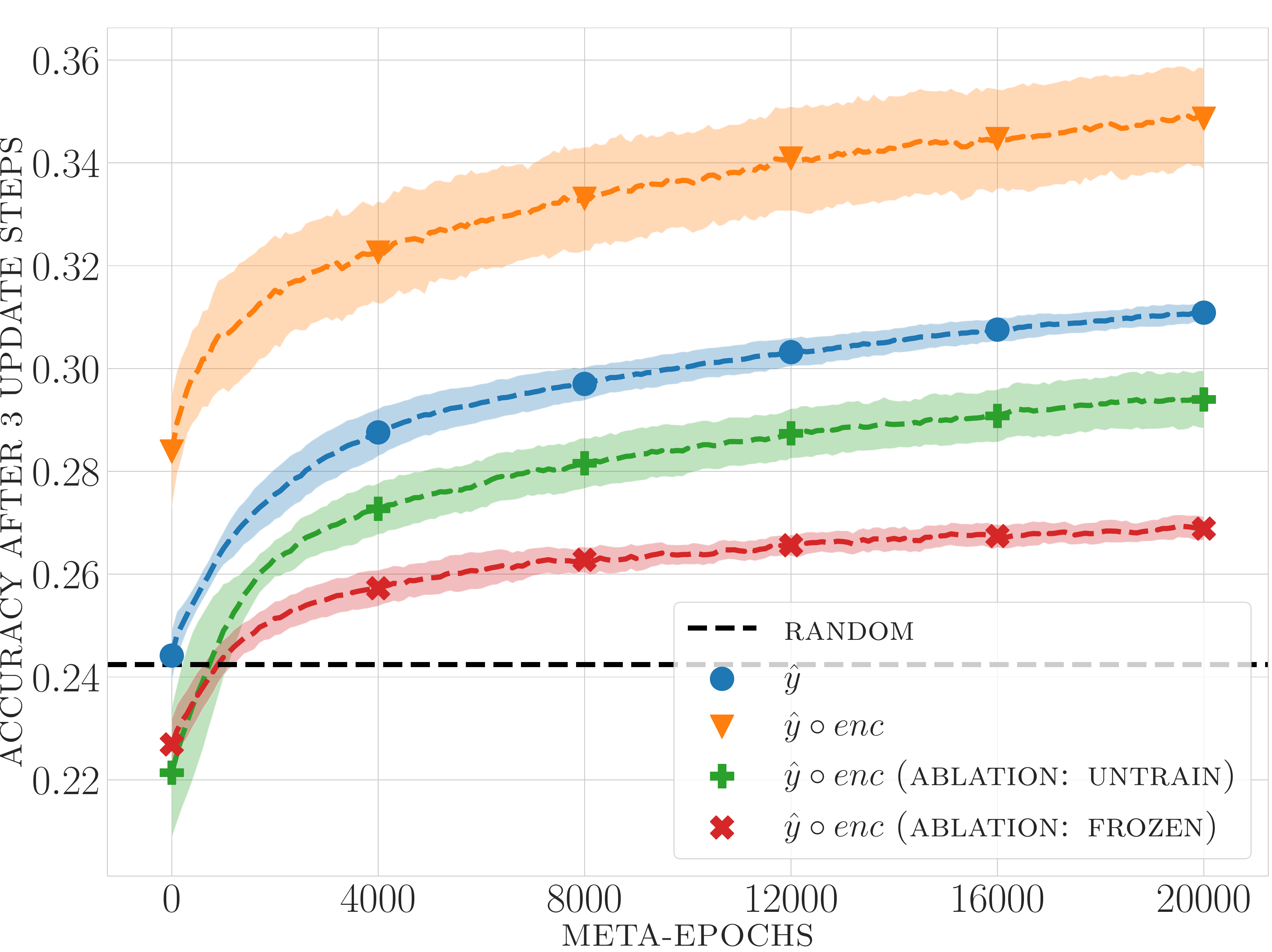}
        \setlength{\belowcaptionskip}{-7pt}
            \caption{\textbf{Latent embedding results}. Meta test accuracy on the \textit{EMNIST-Digits} data set while training on \textit{EMNIST-Letters}. Each point represents the accuracy on the 1600 test tasks after performing three update steps on the respective training~data. The results are averaged over 5 runs and display the mean and standard deviation. 
            }
     \label{winevalzoom}
        \end{minipage}   
        
\end{figure}

\paragraph{Latent Embeddings} Learning to align features is only feasible for unstructured data since this approach would not preserve any structure. Images have an inherent structure in that a permutation of pixels is not meaningful while the channels (for colored inputs) already follow a standard input sequence (red, green, and blue). Likewise, learning to realign pixels across datasets is also unreasonable. Instead, images with varying sizes can be aligned trivially by simply scaling them to the same size.
However, it is a widespread practice among few-shot classification methods, and computer vision approaches in general, to use a pretrained neural network to embed image data into a latent space before applying further operations. The authors \cite{leo} train a Wide Residual Network \citep{zagoruyko2016wide} on the meta-training data of \textit{mini}ImageNet \citep{vinyals2016matching} to compute latent embeddings of the data which are then used for few-shot classification, demonstrating state-of-the-art results. They publicized the embeddings for reproducibility, but not the feature extracting network, making it impossible to apply the trained model to novel tasks that are not embedded yet. We can use \cha{} to align the latent embeddings of image datasets that are generated with different networks. Thus, it is possible to use latent embeddings for meta-training while evaluating on novel tasks that are not yet embedded in case the embedding network is not available, or the complexity of different datasets requires models with different capacities to extract useful features.

For this reason, we conduct an additional experiment for which we combine two similar image datasets, namely \textit{EMNIST-Digits} and \textit{EMNIST-Letters} \citep{cohen2017emnist}. They contain $280,000$ and $145,600$ instances of size $28 \times 28$, and 10 and 26 classes respectively. 
Similarly to the work of \cite{leo}, we train one neural network on each dataset in order to generate similar but different latent embeddings. Afterward, we can sample training tasks from one embedding while evaluating on tasks sampled from the other one.
Both networks used for generating the latent embeddings consist of two convolutional and two dense layers, but the number of neurons per dense layer and filters per convolutional layer is 32 for \textit{EMNIST-Digits} and 64 for \textit{EMNIST-Letters}. In the combined experiments, the full training is performed on the \textit{EMNIST-Letters} dataset, while \textit{EMNIST-Digits} is used for testing. Splitting the features is not necessary as the train, and test features are coming from different datasets. For these experiments, the \cha{} component still has two convolutional layers with 8 and 16 filters, while we use a larger base network with two feed-forward layers with 64 neurons each. 

The results of this experiment are displayed in Figure \ref{winevalzoom}. It shows the accuracy of \textit{EMNIST-Digits} averaged across 5 runs with 1,600 generated tasks per run during the \reptile{} training on \textit{EMNIST-Letters} for the different model variants. Each test task is evaluated by performing 3 update steps on the training samples and measuring the accuracy of its validation data afterward.
One can see that our proposed approach reports a significantly higher accuracy than the \reptile{} baseline after performing three update steps on a task (contribution 4). Moreover, simply adding the \cha{} component does not lead to any improvement. This might be sparked by using a \cha{} component that has a much lower number of parameters than the base network. Only by adding the reordering-training, the model manages to converge to a suitable initialization. In contrast to our experiments on the OpenML datasets, freezing the weights of \cha{} after pretraining also fails to give an improvement, suggesting that the pretraining did not manage to capture the ideal alignment, but enables learning it during joint training.
Our code is available online for reproduction purposes at \textsc{\url{https://github.com/radrumond/chameleon}}.

 \section{Conclusion}\label{conclusion}

In this paper, we presented, to the best of our knowledge, the first approach to tackle few-shot classification for unstructured tasks with different schema. Our model component \cha{} is capable of embedding tasks to a common representation by computing a matrix that can reorder the features. For this, we propose a novel pretraining framework that is shown to learn useful permutations across tasks in a supervised fashion without requiring actual labels. 
In experiments on 23 datasets of the OpenML-CC18 benchmark, our method shows significant improvements even when presented with features not seen during training. Furthermore, by aligning different latent embeddings we demonstrate how a single meta-model can be used to learn across multiple image datasets each embedded with a distinct network. 

As future work, we would like to extend \cha{} to time-series features and reinforcement learning, as these tend to present more variations in different tasks, and they would benefit significantly from our model. Since \cha{} can be used in conjunction with any model and optimization-based meta-learning approach, we would like to analyze how our approach performs and influences with different state-of-the-art base models.

\begin{table}[H]
\centering
\adjustbox{max width=\textwidth}{

\begin{tabular}{cc}
 & Loss \\
\begin{tabular}{c}
Dataset        \\ \midrule
segmen \\
jungle \\
wine \\
wilt \\
cmc \\
electr \\
letter \\
phonem \\
vehicl \\
mfeat \\
ilpd \\
Gestur \\
MagicT \\
tic \\
bankno \\
diabet \\
wdbc \\
blood \\
vowel \\
pendig \\
wall \\
abalon \\
analca \\
\end{tabular}
&
\begin{tabular}{cccccc}
\textsc{random} & $\hat y$ &  $ \hat y \circ \enc$ \textsc{(untrain)} &  $ \hat y \circ \enc$  &  $ \hat y \circ \enc$ \textsc{(frozen)}  &  $\hat y$ \textsc{(oracle)}  \\ \midrule
 2.157 $\rpm$ 0.003  &  1.409 $\rpm$ 0.020  &  1.203 $\rpm$ 0.056  &  0.928 $\rpm$ 0.022  & \textbf{ 0.901 $\rpm$ 0.030}  &  0.940 $\rpm$ 0.030 \\ 
 1.324 $\rpm$ 0.004  &  1.079 $\rpm$ 0.002  &  1.086 $\rpm$ 0.002  &  1.081 $\rpm$ 0.002  & \textbf{ 1.077 $\rpm$ 0.002}  &  1.023 $\rpm$ 0.003 \\ 
 1.851 $\rpm$ 0.005  &  1.580 $\rpm$ 0.003  &  1.567 $\rpm$ 0.002  & \textbf{ 1.506 $\rpm$ 0.006}  &  1.513 $\rpm$ 0.012  &  1.512 $\rpm$ 0.009 \\ 
 0.848 $\rpm$ 0.005  &  0.631 $\rpm$ 0.002  &  0.653 $\rpm$ 0.005  &  0.555 $\rpm$ 0.008  & \textbf{ 0.549 $\rpm$ 0.007}  &  0.541 $\rpm$ 0.004 \\ 
 1.327 $\rpm$ 0.003  &  1.086 $\rpm$ 0.003  &  1.057 $\rpm$ 0.007  & \textbf{ 1.039 $\rpm$ 0.002}  &  1.042 $\rpm$ 0.003  &  1.035 $\rpm$ 0.007 \\ 
 0.869 $\rpm$ 0.004  &  0.686 $\rpm$ 0.004  &  0.683 $\rpm$ 0.002  & \textbf{ 0.639 $\rpm$ 0.007}  &  0.641 $\rpm$ 0.007  &  0.655 $\rpm$ 0.008 \\ 
 3.426 $\rpm$ 0.001  &  3.150 $\rpm$ 0.020  &  3.033 $\rpm$ 0.017  &  2.909 $\rpm$ 0.024  & \textbf{ 2.689 $\rpm$ 0.031}  &  2.377 $\rpm$ 0.031 \\ 
 0.858 $\rpm$ 0.002  & \textbf{ 0.665 $\rpm$ 0.005}  &  0.684 $\rpm$ 0.004  &  0.665 $\rpm$ 0.005  &  0.668 $\rpm$ 0.004  &  0.577 $\rpm$ 0.005 \\ 
 1.624 $\rpm$ 0.004  &  1.310 $\rpm$ 0.020  &  1.227 $\rpm$ 0.008  &  1.214 $\rpm$ 0.033  & \textbf{ 1.199 $\rpm$ 0.037}  &  1.063 $\rpm$ 0.012 \\ 
 2.535 $\rpm$ 0.005  &  1.681 $\rpm$ 0.031  &  1.486 $\rpm$ 0.036  & \textbf{ 1.370 $\rpm$ 0.027}  &  1.405 $\rpm$ 0.026  &  1.359 $\rpm$ 0.049 \\ 
 0.831 $\rpm$ 0.006  &  0.626 $\rpm$ 0.004  &  0.615 $\rpm$ 0.003  & \textbf{ 0.603 $\rpm$ 0.004}  &  0.611 $\rpm$ 0.003  &  0.605 $\rpm$ 0.006 \\ 
 1.809 $\rpm$ 0.002  &  1.499 $\rpm$ 0.006  &  1.437 $\rpm$ 0.006  & \textbf{ 1.419 $\rpm$ 0.002}  &  1.421 $\rpm$ 0.004  &  1.398 $\rpm$ 0.006 \\ 
 0.853 $\rpm$ 0.002  &  0.649 $\rpm$ 0.003  &  0.652 $\rpm$ 0.008  &  0.604 $\rpm$ 0.007  & \textbf{ 0.603 $\rpm$ 0.004}  &  0.590 $\rpm$ 0.007 \\ 
 0.871 $\rpm$ 0.002  &  0.698 $\rpm$ 0.001  &  0.694 $\rpm$ 0.001  & \textbf{ 0.690 $\rpm$ 0.000}  &  0.690 $\rpm$ 0.001  &  0.610 $\rpm$ 0.004 \\ 
 0.840 $\rpm$ 0.009  &  0.639 $\rpm$ 0.004  &  0.654 $\rpm$ 0.006  & \textbf{ 0.616 $\rpm$ 0.001}  &  0.621 $\rpm$ 0.002  &  0.569 $\rpm$ 0.003 \\ 
 0.851 $\rpm$ 0.003  &  0.623 $\rpm$ 0.002  &  0.638 $\rpm$ 0.004  &  0.605 $\rpm$ 0.002  & \textbf{ 0.598 $\rpm$ 0.003}  &  0.600 $\rpm$ 0.004 \\ 
 0.823 $\rpm$ 0.010  &  0.311 $\rpm$ 0.026  &  0.221 $\rpm$ 0.014  & \textbf{ 0.158 $\rpm$ 0.007}  &  0.194 $\rpm$ 0.014  &  0.197 $\rpm$ 0.013 \\ 
 0.845 $\rpm$ 0.004  &  0.681 $\rpm$ 0.003  &  0.688 $\rpm$ 0.002  &  0.660 $\rpm$ 0.003  & \textbf{ 0.659 $\rpm$ 0.002}  &  0.647 $\rpm$ 0.001 \\ 
 2.641 $\rpm$ 0.003  &  2.315 $\rpm$ 0.015  &  1.912 $\rpm$ 0.016  & \textbf{ 1.821 $\rpm$ 0.023}  &  1.843 $\rpm$ 0.021  &  1.671 $\rpm$ 0.029 \\ 
 2.545 $\rpm$ 0.004  &  2.189 $\rpm$ 0.006  &  2.169 $\rpm$ 0.010  &  2.107 $\rpm$ 0.020  & \textbf{ 2.099 $\rpm$ 0.021}  &  1.068 $\rpm$ 0.034 \\ 
 1.638 $\rpm$ 0.003  &  1.360 $\rpm$ 0.011  &  1.083 $\rpm$ 0.014  & \textbf{ 0.972 $\rpm$ 0.014}  &  0.986 $\rpm$ 0.007  &  0.868 $\rpm$ 0.016 \\ 
 1.311 $\rpm$ 0.003  &  0.871 $\rpm$ 0.005  &  0.894 $\rpm$ 0.009  & \textbf{ 0.828 $\rpm$ 0.004}  &  0.834 $\rpm$ 0.005  &  0.823 $\rpm$ 0.007 \\ 
 2.062 $\rpm$ 0.002  &  1.801 $\rpm$ 0.000  & \textbf{ 1.794 $\rpm$ 0.001}  &  1.806 $\rpm$ 0.002  &  1.806 $\rpm$ 0.001  &  1.827 $\rpm$ 0.004 \\ 
\end{tabular} \\ \midrule

 & Accuracy \\
\begin{tabular}{c}
Dataset        \\ \midrule
segmen \\
jungle \\
wine \\
wilt \\
cmc \\
electr \\
letter \\
phonem \\
vehicl \\
mfeat \\
ilpd \\
Gestur \\
MagicT \\
tic \\
bankno \\
diabet \\
wdbc \\
blood \\
vowel \\
pendig \\
wall \\
abalon \\
analca \\
\end{tabular}

&

\begin{tabular}{cccccc}
\textsc{random} & $\hat y$  &  $ \hat y \circ \enc$ \textsc{(untrain)} &  $ \hat y \circ \enc$ &  $ \hat y \circ \enc$ \textsc{(frozen)}  &  $\hat y$ \textsc{(oracle)}  \\ \midrule
 0.147 $\rpm$ 0.001  &  0.419 $\rpm$ 0.005  &  0.496 $\rpm$ 0.015  &  0.595 $\rpm$ 0.012  & \textbf{ 0.619 $\rpm$ 0.015}  &  0.605 $\rpm$ 0.009 \\ 
 0.335 $\rpm$ 0.002  &  0.395 $\rpm$ 0.003  &  0.382 $\rpm$ 0.003  &  0.393 $\rpm$ 0.004  & \textbf{ 0.396 $\rpm$ 0.002}  &  0.460 $\rpm$ 0.003 \\ 
 0.201 $\rpm$ 0.002  &  0.264 $\rpm$ 0.003  &  0.273 $\rpm$ 0.003  & \textbf{ 0.314 $\rpm$ 0.005}  &  0.308 $\rpm$ 0.007  &  0.312 $\rpm$ 0.009 \\ 
 0.504 $\rpm$ 0.002  &  0.628 $\rpm$ 0.002  &  0.601 $\rpm$ 0.009  &  0.718 $\rpm$ 0.005  & \textbf{ 0.720 $\rpm$ 0.005}  &  0.724 $\rpm$ 0.003 \\ 
 0.331 $\rpm$ 0.002  &  0.386 $\rpm$ 0.004  &  0.422 $\rpm$ 0.008  & \textbf{ 0.448 $\rpm$ 0.003}  &  0.446 $\rpm$ 0.002  &  0.461 $\rpm$ 0.008 \\ 
 0.499 $\rpm$ 0.002  &  0.548 $\rpm$ 0.010  &  0.559 $\rpm$ 0.004  & \textbf{ 0.626 $\rpm$ 0.011}  &  0.625 $\rpm$ 0.007  &  0.603 $\rpm$ 0.011 \\ 
 0.039 $\rpm$ 0.000  &  0.078 $\rpm$ 0.005  &  0.112 $\rpm$ 0.005  &  0.153 $\rpm$ 0.006  & \textbf{ 0.204 $\rpm$ 0.006}  &  0.282 $\rpm$ 0.012 \\ 
 0.504 $\rpm$ 0.003  &  0.594 $\rpm$ 0.012  &  0.561 $\rpm$ 0.005  & \textbf{ 0.600 $\rpm$ 0.008}  &  0.597 $\rpm$ 0.008  &  0.702 $\rpm$ 0.001 \\ 
 0.255 $\rpm$ 0.001  &  0.366 $\rpm$ 0.010  &  0.413 $\rpm$ 0.010  &  0.418 $\rpm$ 0.022  & \textbf{ 0.434 $\rpm$ 0.027}  &  0.523 $\rpm$ 0.009 \\ 
 0.104 $\rpm$ 0.002  &  0.354 $\rpm$ 0.006  &  0.398 $\rpm$ 0.008  &  0.428 $\rpm$ 0.012  & \textbf{ 0.431 $\rpm$ 0.009}  &  0.447 $\rpm$ 0.012 \\ 
 0.506 $\rpm$ 0.003  &  0.654 $\rpm$ 0.005  &  0.659 $\rpm$ 0.004  & \textbf{ 0.670 $\rpm$ 0.005}  &  0.662 $\rpm$ 0.006  &  0.669 $\rpm$ 0.006 \\ 
 0.202 $\rpm$ 0.002  &  0.310 $\rpm$ 0.002  &  0.350 $\rpm$ 0.006  & \textbf{ 0.368 $\rpm$ 0.003}  &  0.364 $\rpm$ 0.004  &  0.383 $\rpm$ 0.002 \\ 
 0.503 $\rpm$ 0.002  &  0.611 $\rpm$ 0.002  &  0.601 $\rpm$ 0.012  & \textbf{ 0.662 $\rpm$ 0.007}  &  0.661 $\rpm$ 0.005  &  0.672 $\rpm$ 0.004 \\ 
 0.502 $\rpm$ 0.002  &  0.504 $\rpm$ 0.003  &  0.510 $\rpm$ 0.003  &  0.533 $\rpm$ 0.001  & \textbf{ 0.534 $\rpm$ 0.005}  &  0.666 $\rpm$ 0.005 \\ 
 0.506 $\rpm$ 0.005  & \textbf{ 0.634 $\rpm$ 0.005}  &  0.622 $\rpm$ 0.003  &  0.629 $\rpm$ 0.004  &  0.626 $\rpm$ 0.004  &  0.652 $\rpm$ 0.003 \\ 
 0.505 $\rpm$ 0.004  &  0.656 $\rpm$ 0.004  &  0.639 $\rpm$ 0.007  & \textbf{ 0.674 $\rpm$ 0.002}  &  0.673 $\rpm$ 0.003  &  0.675 $\rpm$ 0.006 \\ 
 0.521 $\rpm$ 0.007  &  0.882 $\rpm$ 0.008  &  0.906 $\rpm$ 0.008  & \textbf{ 0.937 $\rpm$ 0.003}  &  0.918 $\rpm$ 0.007  &  0.919 $\rpm$ 0.006 \\ 
 0.502 $\rpm$ 0.001  &  0.579 $\rpm$ 0.012  &  0.558 $\rpm$ 0.010  &  0.613 $\rpm$ 0.007  & \textbf{ 0.615 $\rpm$ 0.003}  &  0.636 $\rpm$ 0.004 \\ 
 0.092 $\rpm$ 0.001  &  0.143 $\rpm$ 0.007  &  0.303 $\rpm$ 0.007  & \textbf{ 0.346 $\rpm$ 0.010}  &  0.336 $\rpm$ 0.008  &  0.391 $\rpm$ 0.013 \\ 
 0.102 $\rpm$ 0.001  &  0.180 $\rpm$ 0.003  &  0.193 $\rpm$ 0.004  &  0.222 $\rpm$ 0.011  & \textbf{ 0.227 $\rpm$ 0.009}  &  0.646 $\rpm$ 0.010 \\ 
 0.254 $\rpm$ 0.001  &  0.324 $\rpm$ 0.012  &  0.494 $\rpm$ 0.007  & \textbf{ 0.576 $\rpm$ 0.009}  &  0.562 $\rpm$ 0.007  &  0.631 $\rpm$ 0.010 \\ 
 0.339 $\rpm$ 0.003  &  0.566 $\rpm$ 0.002  &  0.554 $\rpm$ 0.007  & \textbf{ 0.594 $\rpm$ 0.004}  &  0.587 $\rpm$ 0.004  &  0.593 $\rpm$ 0.005 \\ 
 0.166 $\rpm$ 0.001  &  0.170 $\rpm$ 0.000  &  0.170 $\rpm$ 0.002  & \textbf{ 0.172 $\rpm$ 0.002}  &  0.171 $\rpm$ 0.002  &  0.179 $\rpm$ 0.002 \\ 
\end{tabular}

\end{tabular}}
\caption{Loss and accuracy scores of each model variant for the \textit{No-Split} experiments. The values depict the mean and standard deviation across 5 runs for each dataset with 1600 sampled test tasks per run. Best results are boldfaced (excluding \textsc{oracle}).}
\label{statsexpNOSPLIT}
\end{table}
\begin{table}[H]
\centering
\adjustbox{max width=\textwidth}{
\begin{tabular}{cc}
 & Loss \\
\begin{tabular}{c}
Dataset        \\ \midrule
vowel \\
wdbc \\
jungle \\
phonem \\
wine \\
analca \\
MagicT \\
diabet \\
letter \\
ilpd \\
Gestur \\
mfeat \\
wilt \\
wall \\
segmen \\
cmc \\
pendig \\
electr \\
vehicl \\
abalon \\
tic \\
\end{tabular}
&

\begin{tabular}{cccccc}
\textsc{random} & $\hat y$ &  $ \hat y \circ \enc$ \textsc{(untrain)} &  $ \hat y \circ \enc$ &  $ \hat y \circ \enc$ \textsc{(frozen)}  \\ \midrule
 2.640 $\rpm$ 0.001  &  2.313 $\rpm$ 0.007  &  1.969 $\rpm$ 0.016  &  1.913 $\rpm$ 0.013  & \textbf{ 1.911 $\rpm$ 0.016} \\ 
 0.826 $\rpm$ 0.014  &  0.264 $\rpm$ 0.031  &  0.167 $\rpm$ 0.010  & \textbf{ 0.162 $\rpm$ 0.002}  &  0.170 $\rpm$ 0.006 \\ 
 1.332 $\rpm$ 0.004  &  1.142 $\rpm$ 0.014  & \textbf{ 1.089 $\rpm$ 0.002}  &  1.099 $\rpm$ 0.005  &  1.093 $\rpm$ 0.004 \\ 
 0.856 $\rpm$ 0.003  &  0.769 $\rpm$ 0.034  &  0.719 $\rpm$ 0.005  &  0.720 $\rpm$ 0.010  & \textbf{ 0.716 $\rpm$ 0.009} \\ 
 1.855 $\rpm$ 0.002  &  1.596 $\rpm$ 0.004  &  1.582 $\rpm$ 0.005  & \textbf{ 1.546 $\rpm$ 0.018}  &  1.547 $\rpm$ 0.013 \\ 
 2.061 $\rpm$ 0.001  &  1.802 $\rpm$ 0.002  & \textbf{ 1.794 $\rpm$ 0.001}  &  1.796 $\rpm$ 0.001  &  1.798 $\rpm$ 0.002 \\ 
 0.851 $\rpm$ 0.004  &  0.673 $\rpm$ 0.009  &  0.662 $\rpm$ 0.002  & \textbf{ 0.629 $\rpm$ 0.007}  &  0.630 $\rpm$ 0.004 \\ 
 0.850 $\rpm$ 0.004  &  0.675 $\rpm$ 0.009  &  0.677 $\rpm$ 0.008  & \textbf{ 0.646 $\rpm$ 0.012}  &  0.655 $\rpm$ 0.014 \\ 
 3.426 $\rpm$ 0.001  &  3.160 $\rpm$ 0.017  &  3.058 $\rpm$ 0.009  &  2.980 $\rpm$ 0.031  & \textbf{ 2.782 $\rpm$ 0.032} \\ 
 0.840 $\rpm$ 0.002  &  0.692 $\rpm$ 0.005  & \textbf{ 0.689 $\rpm$ 0.008}  &  0.694 $\rpm$ 0.004  &  0.694 $\rpm$ 0.004 \\ 
 1.813 $\rpm$ 0.003  &  1.514 $\rpm$ 0.006  &  1.429 $\rpm$ 0.004  & \textbf{ 1.413 $\rpm$ 0.005}  &  1.416 $\rpm$ 0.003 \\ 
 2.531 $\rpm$ 0.005  &  1.591 $\rpm$ 0.067  & \textbf{ 1.417 $\rpm$ 0.010}  &  1.627 $\rpm$ 0.053  &  1.620 $\rpm$ 0.059 \\ 
 0.844 $\rpm$ 0.003  &  0.721 $\rpm$ 0.034  &  0.652 $\rpm$ 0.007  & \textbf{ 0.633 $\rpm$ 0.026}  &  0.671 $\rpm$ 0.019 \\ 
 1.640 $\rpm$ 0.004  &  1.356 $\rpm$ 0.002  &  1.081 $\rpm$ 0.009  & \textbf{ 0.993 $\rpm$ 0.010}  &  1.003 $\rpm$ 0.009 \\ 
 2.166 $\rpm$ 0.002  &  1.388 $\rpm$ 0.061  &  1.147 $\rpm$ 0.021  & \textbf{ 0.799 $\rpm$ 0.024}  &  0.840 $\rpm$ 0.020 \\ 
 1.327 $\rpm$ 0.001  &  1.098 $\rpm$ 0.003  &  1.086 $\rpm$ 0.003  & \textbf{ 1.076 $\rpm$ 0.011}  &  1.082 $\rpm$ 0.004 \\ 
 2.548 $\rpm$ 0.003  &  2.210 $\rpm$ 0.016  &  2.195 $\rpm$ 0.015  &  2.123 $\rpm$ 0.009  & \textbf{ 2.038 $\rpm$ 0.196} \\ 
 0.865 $\rpm$ 0.003  &  0.691 $\rpm$ 0.005  &  0.686 $\rpm$ 0.001  & \textbf{ 0.642 $\rpm$ 0.005}  &  0.646 $\rpm$ 0.007 \\ 
 1.624 $\rpm$ 0.005  &  1.289 $\rpm$ 0.008  &  1.221 $\rpm$ 0.004  & \textbf{ 1.193 $\rpm$ 0.018}  &  1.225 $\rpm$ 0.004 \\ 
 1.313 $\rpm$ 0.004  &  0.971 $\rpm$ 0.025  &  0.929 $\rpm$ 0.004  & \textbf{ 0.894 $\rpm$ 0.014}  &  0.910 $\rpm$ 0.003 \\ 
 0.870 $\rpm$ 0.003  &  0.703 $\rpm$ 0.003  &  0.695 $\rpm$ 0.001  &  0.696 $\rpm$ 0.002  & \textbf{ 0.694 $\rpm$ 0.001} \\ 
\end{tabular} \\ \midrule

 & Accuracy \\
\begin{tabular}{c}
Dataset        \\ \midrule
vowel \\
wdbc \\
jungle \\
phonem \\
wine \\
analca \\
MagicT \\
diabet \\
letter \\
ilpd \\
Gestur \\
mfeat \\
wilt \\
wall \\
segmen \\
cmc \\
pendig \\
electr \\
vehicl \\
abalon \\
tic \\
\end{tabular}
&

\begin{tabular}{ccccc}
\textsc{random} & $\hat y$ &  $ \hat y \circ \enc$ \textsc{(untrain)} &  $ \hat y \circ \enc$ &  $ \hat y \circ \enc$ \textsc{(frozen)}  \\ \midrule
 0.092 $\rpm$ 0.001  &  0.144 $\rpm$ 0.003  &  0.288 $\rpm$ 0.007  & \textbf{ 0.311 $\rpm$ 0.009}  &  0.311 $\rpm$ 0.007 \\ 
 0.522 $\rpm$ 0.009  &  0.901 $\rpm$ 0.011  &  0.937 $\rpm$ 0.004  & \textbf{ 0.942 $\rpm$ 0.003}  &  0.935 $\rpm$ 0.004 \\ 
 0.333 $\rpm$ 0.001  &  0.359 $\rpm$ 0.011  &  0.385 $\rpm$ 0.004  &  0.378 $\rpm$ 0.009  & \textbf{ 0.388 $\rpm$ 0.005} \\ 
 0.503 $\rpm$ 0.002  &  0.504 $\rpm$ 0.021  &  0.502 $\rpm$ 0.017  &  0.529 $\rpm$ 0.004  & \textbf{ 0.533 $\rpm$ 0.024} \\ 
 0.201 $\rpm$ 0.002  &  0.248 $\rpm$ 0.004  &  0.265 $\rpm$ 0.007  & \textbf{ 0.289 $\rpm$ 0.010}  &  0.285 $\rpm$ 0.010 \\ 
 0.167 $\rpm$ 0.001  &  0.172 $\rpm$ 0.003  &  0.173 $\rpm$ 0.002  & \textbf{ 0.185 $\rpm$ 0.002}  &  0.182 $\rpm$ 0.002 \\ 
 0.502 $\rpm$ 0.002  &  0.582 $\rpm$ 0.010  &  0.586 $\rpm$ 0.003  &  0.634 $\rpm$ 0.010  & \textbf{ 0.634 $\rpm$ 0.004} \\ 
 0.501 $\rpm$ 0.002  &  0.601 $\rpm$ 0.012  &  0.605 $\rpm$ 0.008  & \textbf{ 0.635 $\rpm$ 0.012}  &  0.635 $\rpm$ 0.017 \\ 
 0.039 $\rpm$ 0.000  &  0.080 $\rpm$ 0.005  &  0.108 $\rpm$ 0.003  &  0.137 $\rpm$ 0.007  & \textbf{ 0.181 $\rpm$ 0.006} \\ 
 0.501 $\rpm$ 0.002  &  0.571 $\rpm$ 0.004  &  0.579 $\rpm$ 0.003  & \textbf{ 0.583 $\rpm$ 0.006}  &  0.580 $\rpm$ 0.005 \\ 
 0.200 $\rpm$ 0.001  &  0.306 $\rpm$ 0.003  &  0.361 $\rpm$ 0.003  & \textbf{ 0.375 $\rpm$ 0.004}  &  0.372 $\rpm$ 0.004 \\ 
 0.103 $\rpm$ 0.001  &  0.377 $\rpm$ 0.024  & \textbf{ 0.425 $\rpm$ 0.023}  &  0.336 $\rpm$ 0.027  &  0.335 $\rpm$ 0.015 \\ 
 0.504 $\rpm$ 0.004  &  0.563 $\rpm$ 0.043  &  0.598 $\rpm$ 0.011  & \textbf{ 0.643 $\rpm$ 0.034}  &  0.589 $\rpm$ 0.030 \\ 
 0.252 $\rpm$ 0.002  &  0.330 $\rpm$ 0.004  &  0.487 $\rpm$ 0.005  & \textbf{ 0.553 $\rpm$ 0.004}  &  0.543 $\rpm$ 0.005 \\ 
 0.148 $\rpm$ 0.002  &  0.414 $\rpm$ 0.022  &  0.501 $\rpm$ 0.013  & \textbf{ 0.638 $\rpm$ 0.009}  &  0.617 $\rpm$ 0.013 \\ 
 0.333 $\rpm$ 0.001  &  0.371 $\rpm$ 0.004  &  0.394 $\rpm$ 0.006  & \textbf{ 0.415 $\rpm$ 0.013}  &  0.408 $\rpm$ 0.006 \\ 
 0.102 $\rpm$ 0.001  &  0.173 $\rpm$ 0.005  &  0.181 $\rpm$ 0.008  &  0.229 $\rpm$ 0.002  & \textbf{ 0.257 $\rpm$ 0.079} \\ 
 0.500 $\rpm$ 0.002  &  0.545 $\rpm$ 0.006  &  0.551 $\rpm$ 0.002  & \textbf{ 0.622 $\rpm$ 0.007}  &  0.616 $\rpm$ 0.009 \\ 
 0.254 $\rpm$ 0.002  &  0.369 $\rpm$ 0.008  &  0.397 $\rpm$ 0.007  & \textbf{ 0.420 $\rpm$ 0.015}  &  0.397 $\rpm$ 0.010 \\ 
 0.338 $\rpm$ 0.002  &  0.513 $\rpm$ 0.016  &  0.532 $\rpm$ 0.003  & \textbf{ 0.556 $\rpm$ 0.010}  &  0.543 $\rpm$ 0.003 \\ 
 0.501 $\rpm$ 0.002  &  0.501 $\rpm$ 0.001  &  0.506 $\rpm$ 0.003  &  0.514 $\rpm$ 0.002  & \textbf{ 0.515 $\rpm$ 0.003} \\ 
\end{tabular}
\end{tabular}}
\caption{Loss and accuracy scores of each model variant for the \textit{Split} experiments. The values depict the mean and standard deviation across 5 runs for each dataset with 1600 sampled test tasks per run. Best results are boldfaced.}
\label{statsexpSPLIT} 
\end{table}

\bibliographystyle{model5-names} 
\biboptions{authoryear}
\bibliography{bibli.bib}

\end{document}